\documentclass{article}
\pdfoutput=1

     \usepackage[preprint,nonatbib]{neurips_2020}

\usepackage[utf8]{inputenc} 
\usepackage[T1]{fontenc}    
\usepackage{url}            
\usepackage{booktabs}       
\usepackage{amsfonts}       
\usepackage{nicefrac}       
\usepackage{microtype}      
\usepackage[dvipsnames]{xcolor}
\usepackage{amsmath}

\usepackage{graphicx}
\usepackage{caption}
\usepackage{subcaption}
\usepackage{bbold}

\definecolor{rangreen}{rgb}{0,0.39,0}

\title{Understand and Improve Contrastive Learning Methods for Visual Representation: A Review}

\author{
  Ran Liu \\
  Machine Learning program - ECE\\
  Georgia Tech\\
  \texttt{rliu361@gatech.edu} \\
}

\begin{document}

\maketitle

\section{Introduction}

Traditional supervised learning methods are hitting a bottleneck \cite{liu2020self} because of their dependency on expensive manually labeled data and their weaknesses such as limited generalization ability and vulnerability to adversarial attacks \cite{goodfellow2014explaining,xu2020neural,liu2020self}. A promising
alternative, self-supervised learning,
as a type of unsupervised learning, has gained popularity because of its potential to learn effective data representations without manual labeling. Among self-supervised learning algorithms, contrastive learning has achieved state-of-the-art performance in several fields of research, including but not limited to computer vision \cite{chen2020simple,he2020momentum, sermanet2018time,zhai2019s4l}, natural language processing \cite{kong2019mutual,lan2019albert}, and biomedical image processing \cite{chen2019self,bai2019self,cheng2020subject,azabou2021mine}.

This literature review aims to provide an up-to-date analysis of researchers' efforts to understand the key components and the limitations of self-supervised learning.
Section \ref{sec2-background} explains a few basic concepts of self-supervied learning.
Section \ref{sec3-PIRL} discusses how pretext tasks are used in self-supervised learning, and how being invariant to image transformations is better than being covariant to image transformations \cite{misra2020self}.
Then, in section \ref{sec4.1-simclr}, we describe a recent contrastive learning method SimCLR \cite{chen2020simple} and in section \ref{sec4.2-vrm} we introduce attempts to improve contrastive methods by altering its key components \cite{kalantidis2020hard}.
Section \ref{sec5.1-dependent} presents the model bias \cite{tian2020makes} in contrastive learning, and is followed by section \ref{sec5.2-general} which discusses possible methods to reduce model bias \cite{wang2020understanding,xiao2020should} in current self-supervised learning methods.

\section{Background}
\label{sec2-background}

\paragraph{Self-supervised learning (SSL)}
Unsupervised learning methods do not use labels from downstream tasks. One type of unsupervised learning methods, self-supervised learning methods, create 'labels' for free for unlabelled data and train unsupervised dataset in a supervised manner, e.g. when doing data augmentation, we can obtain a rotation degree as the pseudolabel when rotating an image.

Self-supervised learning methods can be categorized as {\em generative} approaches and {\em discriminative} approaches. While generative methods require a mapping from image space to feature space $\mathcal{X} \rightarrow \mathcal{Z}$, as well as a mapping from feature space to image space $\mathcal{Z} \rightarrow \mathcal{X}$, discriminative methods only require a mapping from image space to feature space $\mathcal{X} \rightarrow \mathcal{Z}$. Thus, although generative methods have more functionalities, such as constructing images, discriminative methods have certain advantages like requiring fewer parameters in the models. Recent successes of  self-supervised learning methods are built upon discriminative methods, and therefore we will focus on them.

\paragraph{Pretext tasks}
Self-supervised representation learning methods are typically based on pretext tasks. Pretext tasks are pre-designed for networks to solve and the network learns representation based on the training that is regularized by the objective function provided by the pretext tasks. Different from downstream tasks that the network wants to solve in the final stage, pretext tasks do not require manual labeling but has its own semantic meaningful pseudolabels that comes with the definition of the pretext tasks, e.g. when rotating an image we get the rotation angle as the pseudolabel.

Typically, discriminative SSL pretext tasks rely on data augmentations/transformations. In computer vision domain,
classic effective spatial transformation sets include image colorization \cite{dosovitskiy2014discriminative,larsson2016learning}, image affine transformations e.g. rotations \cite{gidaris2018unsupervised,zhang2019aet}, and other geometric transformations e.g. jigsaw transformations \cite{dosovitskiy2014discriminative,noroozi2016unsupervised}.
In video domain, there are also temporal transformation sets, including sorting sequences \cite{lee2017unsupervised,xu2019self,kim2019self}, future prediction \cite{srivastava2015unsupervised,oord2018representation,lorre2020temporal}, and tracking \cite{wang2015unsupervised}. A detailed study of which data augmentation works the best for video self-supervised learning can be found in \cite{qian2020spatiotemporal}.

\paragraph{Contrastive learning}
Contrastive learning is a type of self-supervised representation learning method that is based on recognizing the similarities and differences between objects. The goal of contrastive learning is to group positive examples (similar samples) closer and separate negative examples (dissimilar samples), which can be treated as a specific type of pretext tasks \cite{he2020momentum}.

Typical contrastive learning methods are \emph{instance-specific}, in the sense that positive examples are constructed such that they are different augmented ``views'' of the same sample. Mathematically speaking, given an image $\mathbf{x}$ and a set of possible transformations $\mathcal{T}$, two transformations are sampled as $t_1, t_2 \sim \mathcal{T}$ and $\mathbf{x}$ is augmented as $t_1(\mathbf{x}), t_2(\mathbf{x})$. The resulting $t_1(\mathbf{x})$ and $t_2(\mathbf{x})$ are considered as positive examples to each other because they originated from the same image. All other images are considered as negative examples, regardless of their ground truth labels for downstream tasks (which we don't have access to during training).
We will later mention that this instance-specific method is proven to be effective by several works \cite{misra2020self,chen2020simple,grill2020bootstrap,chen2020exploring}, while is also criticized by several works \cite{xiao2020should}.

\section{Pretext-Covariant Learning or Pretext-Invariant Learning?}
\label{sec3-PIRL}




In this section, we will introduce how exactly SSL methods train models on unsupervised datasets in a supervised manner. We will discuss two ways of using pretext tasks to train the model -- pretext-covariant learning methods and pretext-invariant learning methods. The discussion will be based on the work $\dagger$ Pretext-Invariant Representation Learning (PIRL) \cite{misra2020self} \footnote{PIRL is the main artifact in this section.}.

\paragraph{Pretext-covariant learning}

Most of the earlier SSL methods are based on pretext-covariant learning \cite{doersch2015unsupervised,zhang2016colorful,noroozi2016unsupervised,gidaris2018unsupervised}. Pretext-covariant learning methods aim to learn a representation that is covariant with certain properties of the pretext task's transformations (as mentioned in section \ref{sec2-background}). For example, \cite{gidaris2018unsupervised} proposed a pretext task that predicts image rotation, where the images are augmented by rotating in certain degrees and the network is trained to distinguish the rotation angles. Thus, the network learns a representation that covaries with the rotation angle, and this is pretext covariant representation learning.

Mathematically, given an augmentation set $\mathcal{T}=\{t\mid t \in \mathcal{T}\}$ and an image set $\mathcal{D} = \{\mathbf{x} \mid \mathbf{x} \in \mathcal{D}\}$, pretext covariant learning methods produce a transformed dataset $\mathcal{D}_T  = \{t(\mathbf{x})\mid t \in \mathcal{T}, \mathbf{x} \in \mathcal{D}\}$ with its label being some properties of $t$ as $p(t)$. An encoder $f_{\theta}(\cdot)$ with parameters $\theta$ is trained to extract the representations of images $t(\mathbf{x})$ via $\mathbf{z}_{t(\mathbf{x})}=f_{\theta}(t(\mathbf{x}))$ by minimizing the empirical risk:
\begin{equation}
\label{eq-pretext-covariant}
    \ell_{co}(\theta ; \mathcal{D}) = \mathbb{E}_{t \sim \mathcal{T}}\left[\frac{1}{|\mathcal{D}|} \sum_{\mathbf{x} \in \mathcal{D}} L \left(\mathbf{z}_{t(\mathbf{x})}, p(t)\right)\right]
\end{equation}
where the function $L(\cdot, \cdot)$ measures the mutual information between two inputs, e.g. in the case of rotation $L(\cdot, \cdot)$ would be an operator that predicts the degree of rotation.

\paragraph{Pretext-invariant learning}

\begin{figure*}
\centering
     \begin{subfigure}[b]{0.49\textwidth}
         \centering
         \includegraphics[width=\textwidth]{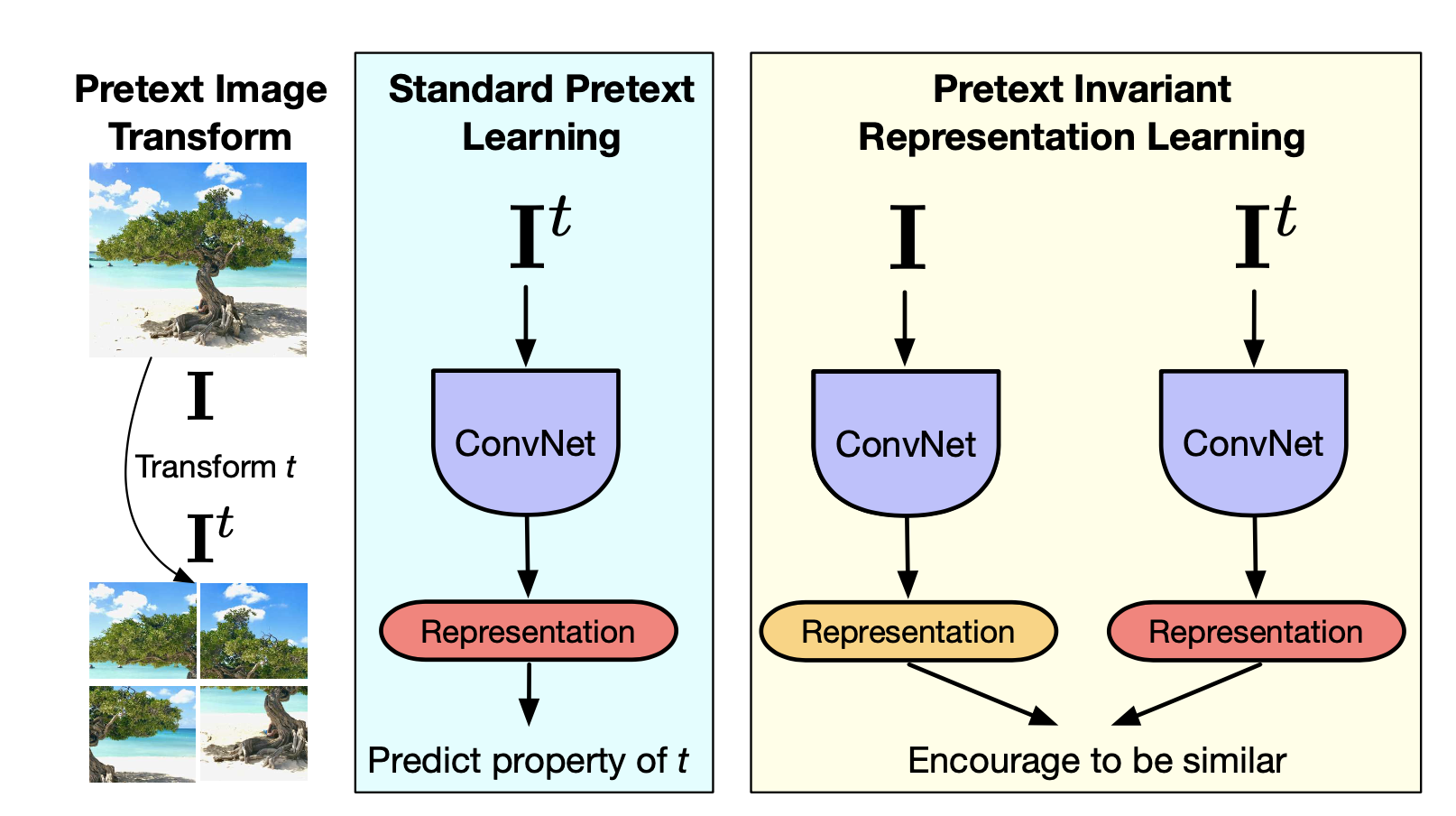}
         \subcaption{}
     \end{subfigure}
     \hfill
     \begin{subfigure}[b]{0.49\textwidth}
         \centering
         \includegraphics[width=\textwidth]{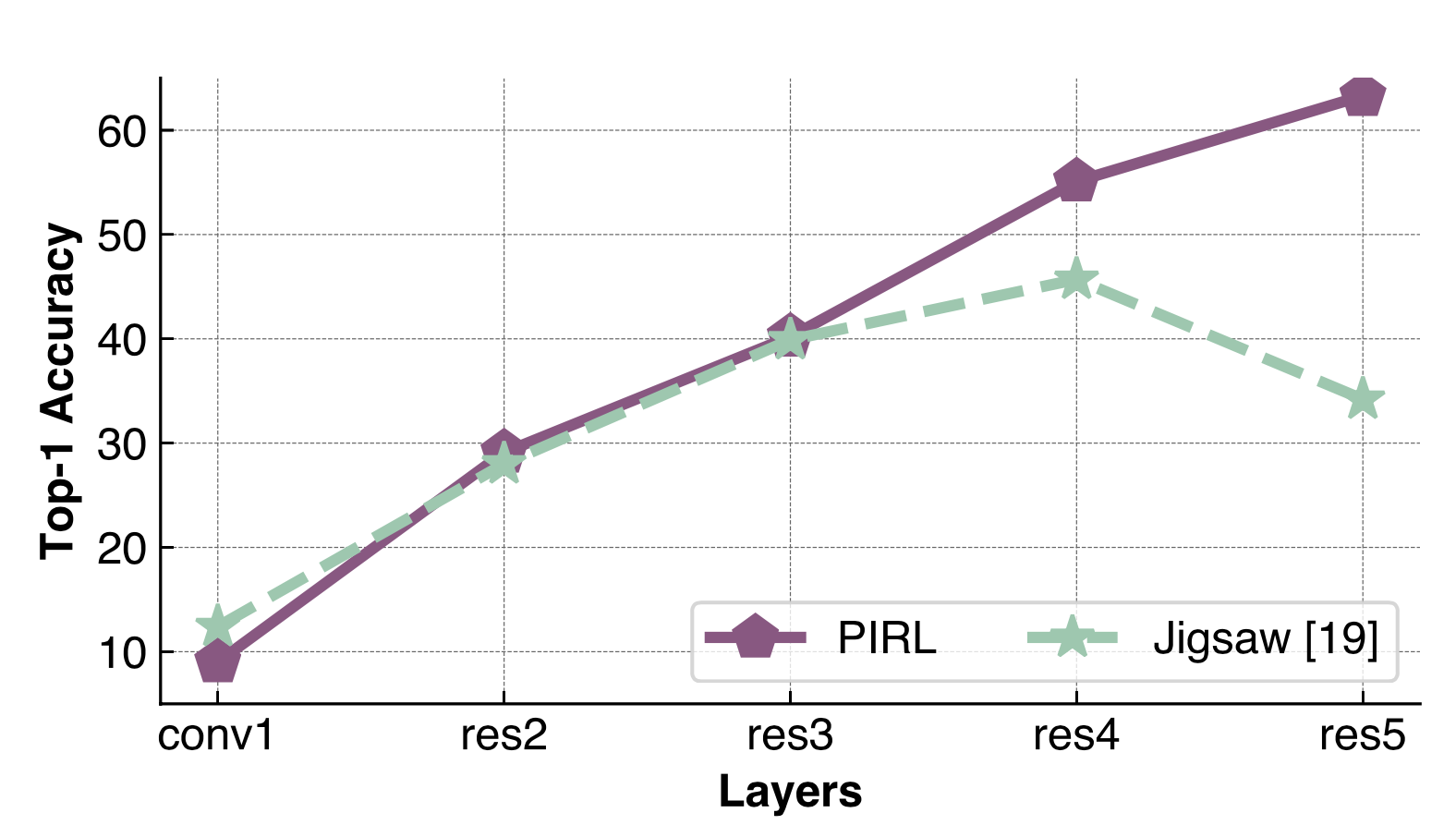}
         \subcaption{}
     \end{subfigure}
    \caption{Figures from the work PIRL \cite{misra2020self}. The LHS figure compares the differences between pretext covariant networks (as standard pretext learning) and pretext invariant networks. The RHS figure shows statistics that support the argument that pretext-invariant learning methods outperform pretext-covariant learning methods.}
    \label{fig-pirl}
\end{figure*}

PIRL \cite{misra2020self} is one of the works \cite{bachman2019learning,he2020momentum} that questioned the intuition behind pretext covariant representation learning. The authors argued that representations ought to be invariant under image transformations because the visual semantics remain the same under most of the image transformations \cite{dalal2005histograms,ji2019invariant}, e.g. a rotated image of dog still contains a dog. Thus, a representation that is invariant to image transformations that do not alter visual semantics should be robust.
The spirit of PIRL is summarized in Fig. \ref{fig-pirl} (a), where the standard pretext covariant learning methods and the pretext invariant learning methods are compared. Instead of predicting the representation to be representative of the property of the transformation, PIRL learns its representation to be representative of the visual semantics that are invariant of the transformation.

Mathematically, given the same set of augmentations $\mathcal{T}$, the same image set $\mathcal{D}$, and a same encoder $f_{\theta}(\cdot)$ that extracts representations as in equation \ref{eq-pretext-covariant},
PIRL trains the encoder by minimizing this alternative empirical risk:
\begin{equation}
\label{eq-pretext-invariant}
    \ell_{i n v}(\theta ; \mathcal{D}) = \mathbb{E}_{t \sim \mathcal{T}}\left[\frac{1}{|\mathcal{D}|} \sum_{\mathbf{x} \in \mathcal{D}} L\left(\mathbf{z}_{\mathbf{x}}, \mathbf{z}_{t(\mathbf{x})}\right)\right]
\end{equation}
where the function $L(\cdot, \cdot)$ measures the similarity between two image representations. This $L(\cdot, \cdot)$ encourages the network $f_{\theta}(\cdot)$ to produce the same representation for image $\mathbf{x}$ as for its transformed counterpart $t(\mathbf{x})$ instead of
predicting the properties of $t$ according to $t(\mathbf{x})$. 

PIRL \cite{misra2020self} conducted experiments to show that learning to be invariant, instead of covariant, to augmentation set $\mathcal{T}$ typically would result in better representations, as illustrated in Fig \ref{fig-pirl} (b). In previous transformation-covariant networks, a common observation is that the representation in the final layer of the network is worse than the representation in the penultimate layer, while PIRL does not have this issue. It is reasonable to surmise that since the last layer representation would interact with the loss function directly, the loss function would discourage the last layer to contain  general information (e.g. a loss function that predicts rotation angle would encourage the last layer to contain rotation-only information). Thus, invariant networks typically would learn a better representation than the covariant networks. Due to this observation, works later on typically choose to learn representation that is invariant with augmentations.

\section{Contrastive Learning}

In this section, we will introduce a representative visual self-supervised learning method that achieves supervised learning performance: $\dagger$ a simple framework for contrastive learning of visual representations (SimCLR) \cite{chen2020simple} \footnote{SimCLR is the main artifact in this subsection.}. After that, we will discuss a few vicinal risk minimization (VRM) techniques that improve contrastive learning performance.

\subsection{Contrastive learning method SimCLR}
\label{sec4.1-simclr}

\begin{figure*}
\centering
    \includegraphics[width=0.95\textwidth]{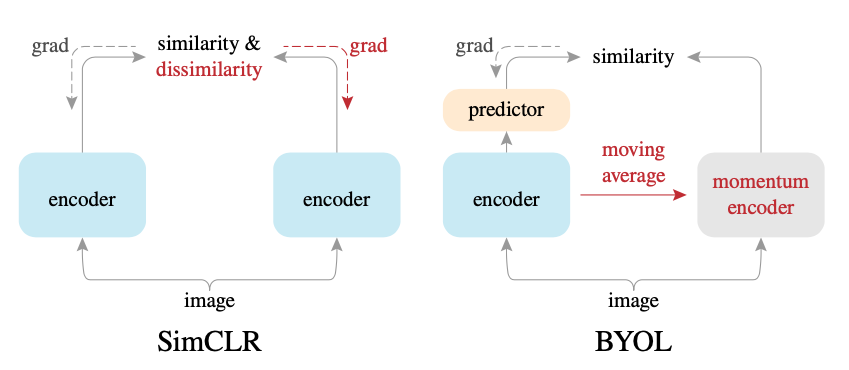}
    \caption{Figures from work SimSiam \cite{chen2020exploring}. The LHS figure shows the architecture of the method SimCLR \cite{chen2020simple}; the RHS figure shows the architecture of the method BYOL \cite{grill2020bootstrap}.}
    \label{fig-simclrbyol}
\end{figure*}

Section \ref{sec3-PIRL} discusses that it is often beneficial to build representation that is invariant to augmentations. SimCLR is one of the representative contrastive learning methods that are built upon invariant representations. Below, we present the method SimCLR, its evaluation, and its key components.

\paragraph{Method} As shown in Fig \ref{fig-simclrbyol}, SimCLR is based on a Siamese-like network that learns to congregate different augmented versions of the same image, and separate dissimilar images (negative examples) to learn latent representations. 
Given a batch of $N$ images, all images are randomly augmented twice to produce $2 N$ augmented data points
\cite{khosla2020supervised}. Thus, for any image $\mathbf{x}$ in the mini-batch, it has $1$ positive example and $2(N-1)$ negative examples inside a mini-batch. An encoder $f_{\theta}$ together with a projector $g_{\theta}$ are used to extract the latent vector that is fed into the loss function, where the encoder representation is $\mathbf{z} = f_\theta(\mathbf{x})$, and the projector representation is $\mathbf{h} = g_\theta(\mathbf{z})$.
Then the loss function for a positive pair of examples $(t_i(\mathbf{x}), t_j(\mathbf{x}))$ is defined as:
\begin{equation}
\label{eq-simclr-loss}
    \ell_{t_i, t_j}=-\log \frac{\exp \left(\operatorname{sim}\left(\boldsymbol{h}_{i}, \boldsymbol{h}_{j}\right) / \tau\right)}{\sum_{k=1}^{2 N} \mathbb{1}_{[k \neq i]} \exp \left(\operatorname{sim}\left(\boldsymbol{h}_{i}, \boldsymbol{h}_{k}\right) / \tau\right)}, \text{where } \operatorname{sim}(\boldsymbol{u}, \boldsymbol{v}) = \frac{\boldsymbol{u}^{\top} \boldsymbol{v}}{\|\boldsymbol{u}\|\|\boldsymbol{v}\|} \text{ for } \boldsymbol{u}, \boldsymbol{v}
\end{equation}
where $\operatorname{sim}(\cdot, \cdot)$ is the cosine similarity, $\mathbb{1}_{[k \neq i]} \in\{0,1\}$ is an indicator function that equals 1 iff $k \neq i$, and $\tau$ denotes a temperature parameter. The final loss for a mini-batch can be obtained by computing and summing up the loss function of all $2 N$ positive pairs (symmetrically) in this mini-batch. SimCLR paper denotes this loss as NT-$\mathrm{Xent}$ (the normalized temperature-scaled cross entropy loss), and meanwhile loss functions of the same format with small variances are also referred to as NCE (noise-contrastive estimation) \cite{sohn2016improved,wu2018unsupervised,hjelm2018learning}  loss and InfoNCE \cite{he2020momentum,oord2018representation} loss.
Intuitively, this contrastive loss can be viewed as a softmax-based classification loss \cite{he2020momentum,wu2018unsupervised} which aims to classify samples from a mini-batch as positive examples and negative examples.

\paragraph{Evaluation}

Intuitively, a model trained in self-supervised manner can be used in two situations. First, when we have no labels at all, the well-trained model should learn good representation with its fixed/frozen weights. Also, when we have limited amount of labels, we want to utilize the available labels and thus fine-tune the pre-trained models with existing labels.

Based on this intuition, the evaluation of self-supervised learning methods typically consists of 1) \emph{linear classifier evaluation}: a linear classifier is trained on top of the frozen representation obtained by the self-supervised training; 2) \emph{semi-supervised learning evaluation}: the self-supervised model is fine-tuned with a small subset of labelled training set using label information; 3) \emph{transfer learning evaluation}: the representation obtained by the self-supervised training on one dataset is either frozen or fine-tuned, and the representation is used either to perform image classification on another dataset, or to solve a different task e.g. object detection.

SimCLR achieved linear evaluation accuracy that matches the performance of a supervised network, and transfer learning accuracy that outperforms the supervised baseline on several datasets. With several improvements including a bigger and deeper projector head, SimCLRv2 \cite{chen2020big} achieved even better performance that outperforms the semi-supervised learning baselines.

\paragraph{Key components}

In SimCLR, the authors conducted a set of ablation experiments to examine the key components of the success of the learning, as we summarized below. 1) The specific composition of challenging data augmentation operations is crucial to learning good representations. 
Otherwise, if an 'easier' set of data augmentation is used, the model could almost perfectly identify the positive pairs in the contrastive task while still learn a bad representation. 2) A larger batch size with more negative examples would significantly benefit the training. This observation is in line with other SSL papers \cite{gutmann2010noise,oord2018representation,wu2018unsupervised} using contrastive loss and the analysis of contrastive loss. 3) The hidden layer before the projection head is a better representation than the layer after, which means that $\mathbf{z}$ is a better representation than $\mathbf{h}$. It is conjectured that this phenomenon is due to the information loss induced by the contrastive loss, as in pretext covariant networks, the penultimate layer representation is always better than the final layer representations. The above three key components of the SimCLR together helped SimCLR and other SSL methods to achieve supervised learning level performances.

\paragraph{Other methods} Aside from SimCLR, there are a lot of other powerful self-supervised learning methods. Here we briefly mention a few other representative and powerful SSL methods.

Similar to SimCLR, MoCo \cite{he2020momentum,chen2020improved} is a SSL method that uses InfoNCE loss. The difference is that while SimCLR uses a large mini-batch size to store negative examples, MoCo stored negative examples using a memory bank, which is the representation computed via a momentum updated version (or moving average version) of the network.

Some other methods such as BYOL \cite{grill2020bootstrap}, SwAV \cite{caron2020unsupervised}, and SimSiam \cite{chen2020exploring} are recently developed to learn self-supervised representation using only positive examples. For example, BYOL (as shown in Fig \ref{fig-simclrbyol}) used a momentum encoder that prevents collapse, and a predictor that induces asymmetry to cluster positive examples.

\subsection{Improve contrastive learning with general vicinal risk minimization}
\label{sec4.2-vrm}

\paragraph{Vicinal risk minimization (VRM)}

VRM \cite{chapelle2001vicinal} is a training principle that draws additional \emph{virtual} examples from the vicinity distribution (neighborhood around each example in the training data) of the training set to enlarge the training distribution. Data augmentation is a common type of VRM method. However, it is not a general VRM method in the sense that the design of data augmentation requires expert knowledge and is dataset dependent.

In the supervised learning field and semi-supervised learning field, several general VRM methods are proposed to improve the training of the deep neural networks. For example, Mixup \cite{zhang2017mixup} and its variants \cite{yun2019cutmix} are proposed to use image mixture to create virtual examples; Manifold Mixup \cite{verma2019manifold} is proposed to use latent representation mixture to create virtual examples.


\paragraph{Hard negative mixing}

\begin{figure}
     \centering
     \begin{subfigure}[b]{0.54\textwidth}
         \centering
         \includegraphics[width=\textwidth]{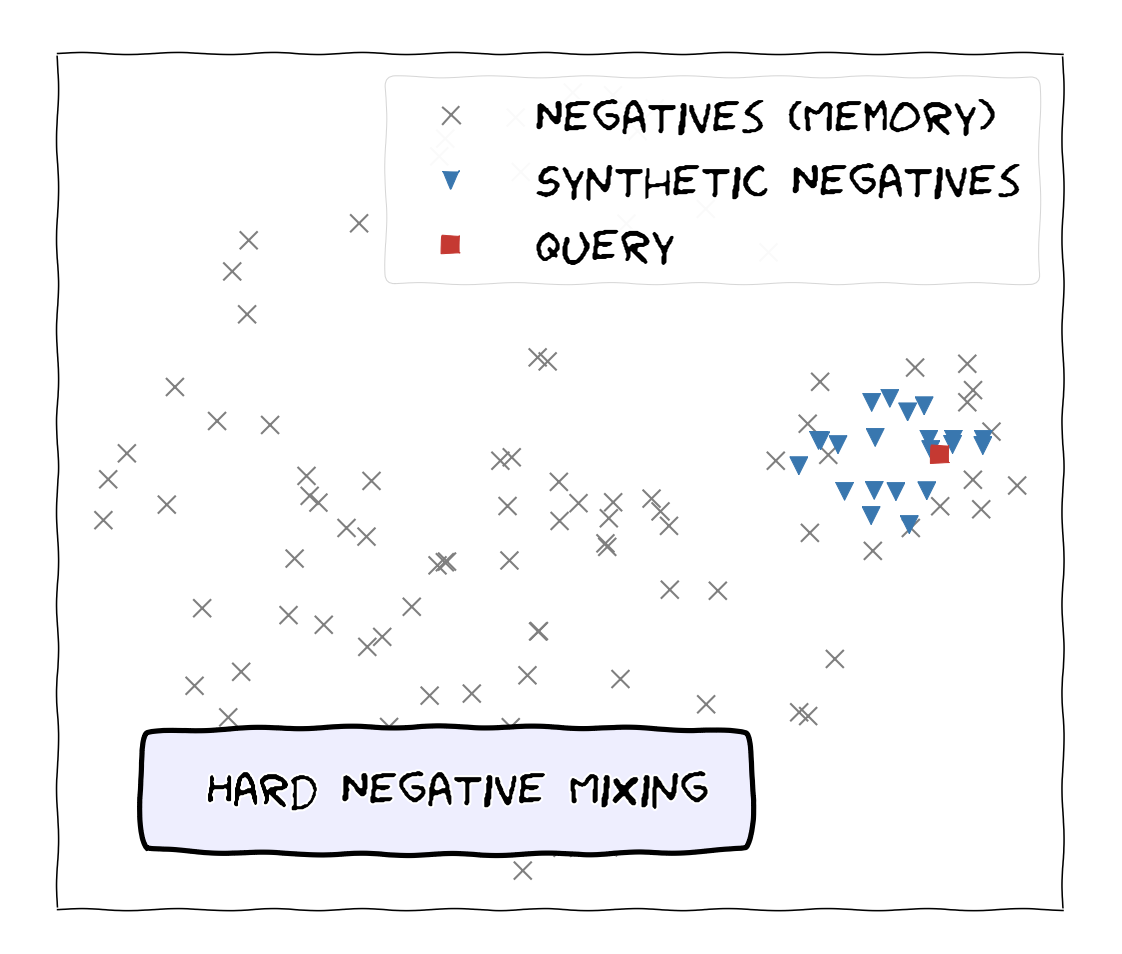}
         \subcaption{}
     \end{subfigure}
     \hfill
     \begin{subfigure}[b]{0.44\textwidth}
         \centering
         \includegraphics[width=\textwidth]{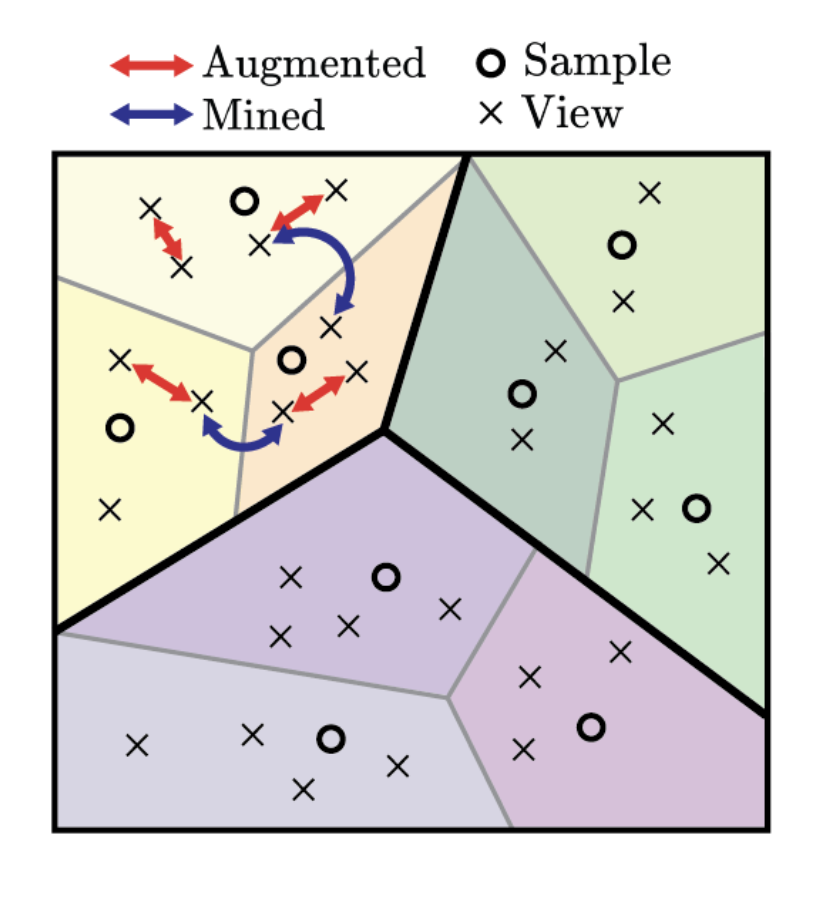}
         \subcaption{}
     \end{subfigure}
    \caption{Latent space manipulation methods. LHS figure is from MoCHi \cite{kalantidis2020hard}, where they synthesize hard negative examples near the data point (query) for MoCo. RHS figure is from MYOW \cite{azabou2021mine}, where they build across-sample views for BYOL.}
    \label{fig: mochi-and-myow}
\end{figure}

As we pointed out in Sec. \ref{sec4.1-simclr}, one of the key components of contrastive methods like SimCLR is the large numbers of the negative examples, i.e. SimCLR used a large mini-batch size and MoCo used a large memory bank to maintain. While the proposed methods in SimCLR and MoCo achieved good performance, the improvement of the performance diminishes with the increase in the amount of negative examples. Why does this happen? $\dagger$ Mixing of contrastive hard negatives (MoCHi) \cite{kalantidis2020hard} \footnote{MoCHi is the main artifact in this subsection.} investigated into this phenomenon and showed that the negative examples are not utilized fully. Thus, MoCHi provided a latent space mixing strategy to increase the contribution of the negative examples and thus improve the performance of contrastive learning method.

Fig.\ref{fig: mochi-and-myow} (a) shows a typical latent space of the contrastive learning method MoCo, where we can see that a lot of negative examples are so far away from the query that they contribute little to the loss function. In other words, the negative examples are too 'easy' and we would need more ``hard negatives'' to make the contrastive methods more effective. The negative examples that are harder to  distinguish can better improve the algorithm performance and thus by generating hard negatives, the contrastive network can learn from fewer examples.

Following this intuition, MoCHi selects and generates negative examples differently, as it generates two types of ``hard negatives'': For any data point $\mathbf{x} \in \mathcal{D}$, we can generate two views $t_i(\mathbf{x})$ and $t_j(\mathbf{x})$ by applying different transformations to the same sample. Let $\mathbf{z}_i$ be the data representation and let $\mathbf{z}_j$ be the positive example representation, and denote all negative example representations as $\mathbf{z}_n \in \mathcal{Q}$, where $\mathcal{Q}$ has a size $K$. According to Eq. \ref{eq-simclr-loss}, the contrastive loss is composed of $K$ negative logits $l\left(\mathbf{z}_{n_k}\right) = \operatorname{sim}(\mathbf{z}_i, \mathbf{z}_{n_k}) / \tau$ and one positive logit $l\left(\mathbf{z}_j\right)=\operatorname{sim}(\mathbf{z}_i, \mathbf{z}_j) / \tau$. MoCHi orders the set of negative features as $\tilde{Q}=\left\{\mathbf{z}_{n_1}, \ldots, \mathbf{z}_{n_K}\right\}$, such that the ordered set obeys: $l\left(\mathbf{z}_{n_i}\right)>l\left(\mathbf{z}_{n_j}\right), \forall i<j,$, i.e. the set is ordered via the negative examples' contribution to the contrastive loss. After ordering the examples, the list is truncated to the $N$ hardest negative examples.
Denote the new set as $\tilde{Q}^{N}$, the authors propose to generate $s$ hard negatives and $s^{\prime}$ harder negatives by creating convex linear combinations between $\mathbf{z}_{n_i}, \mathbf{z}_{n_j} \in \tilde{Q}^{N}$, and between $\mathbf{z}_i$ and  $\mathbf{z}_{n_i} \in \tilde{Q}^{N}$, respectively. Mathematically, the synthetic point $\mathbf{q}_{k}$ and $\mathbf{q}_{k}^{\prime}$ are given by:
\begin{equation}
\begin{aligned}
    s\text{ hard negatives:} \quad & \mathbf{q}_{k}=\frac{\tilde{\mathbf{q}}_{k}}{\left\|\tilde{\mathbf{q}}_{k}\right\|_{2}}, \text { where } \tilde{\mathbf{q}}_{k}=\alpha_{k} \mathbf{z}_{n_i}+\left(1-\alpha_{k}\right) \mathbf{z}_{n_j} \\
    s^{\prime}\text{ harder negatives:} \quad & \mathbf{q}_{k}^{\prime}=\frac{\tilde{\mathbf{q}}_{k}^{\prime}}{\left\|\tilde{\mathbf{q}}_{k}^{\prime}\right\|_{2}}, \text { where } \tilde{\mathbf{q}}_{k}^{\prime}=\beta_{k} \mathbf{z}_i+\left(1-\beta_{k}\right) \mathbf{z}_{n_j}
\end{aligned}
\end{equation}
where $\mathbf{z}_{n_i}, \mathbf{z}_{n_j} \in \tilde{Q}^{N}$ are randomly chosen negative features from the top $N$ existing hard negatives. $\alpha_{k}\in(0,1)$ and $\beta_{k}\in(0,0.5)$ are two randomly chosen mixing coefficient, where $\beta_{k}<0.5$ guarantees that the mixed negative is always more negative than positive. After mixing, the generated logits $l\left(\mathbf{q}_{k}\right)$ and $l\left(\mathbf{q}_{k}^{\prime}\right)$ are computed and appended as additional negative logits.

Using those two latent mixing tricks, MoCHi further increased the performance of contrastive learning methods. More importantly, it provided us a better understanding of the representation space of contrastive learning methods.

\paragraph{Other methods} Aside from MoChi, there are other VRM methods that improve the performance of SSL methods. CLAE \cite{ho2020contrastive} proposed a way to generate challenging positive and negative examples for contrastive learning using an adversarial training algorithm. Un-mix \cite{shen2020rethinking} used image mixtures instead of representation mixtures to improve the quality of the learnt representation.

Another work that worth mentioning is MYOW \cite{azabou2021mine} (as shown in Fig.\ref{fig: mochi-and-myow} (b)), which is based on BYOL so it uses positives examples only. MYOW did not use VRM methods to improve performance, but it shares interesting similarity with MoCHi: as MoCHi shows that negative examples that are close to the data (query) are the key to improve performance, MYOW shows that positive examples that are far away from the data (query) are important to improve performance.

\section{Do Contrastive Learning Methods Learn General Representation?}

There are also works that focus on the model bias and model interpretability of contrastive methods. One research focus is on the generalization ability of the learned features from contrastive methods: Are the current evaluations of SSL methods independent from downstream tasks? How to control the contrastive learning methods such that they can learn more general representation?

\subsection{Contrastive learning methods are still downstream-task dependent}
\label{sec5.1-dependent}

As we mentioned in section \ref{sec4.1-simclr}, the successful learning of the representation is based on a specific composition of data transformation operations. However, an implicit assumption behind the selection of data augmentations is that the selected augmentations are known to keep task-relevant information intact e.g. a rotated image of dog is known to still be an image of dog. However, if the downstream task is completely unknown, we would not know if the augmentation is helpful or not,
e.g. a rotated image of dog would be not a good image to predict the direction of the gravity.

The above intuition is presented by Tian, et al. \cite{tian2020makes} as a proposition, which is defined and illustrated mathematically using the concept of mutual information. Based on this proposition, they studied how different choices of the augmented views affect the quality of the learned representation. They concluded that the current contrastive learning is powerful because they reduce the mutual information (MI) between views while keeping task-relevant information intact. Moreover, they proposed a semi-supervised method to learn effective views for a known given task while keeping the SSL process unsupervised.

This analysis revealed a crucial limitation of current SSL methods: the successes of SSL are still based on ``supervision'', which is implicitly introduced during the selection of data transformations. The specific composition of data augmentation operations (in section \ref{sec4.1-simclr}) is carefully designed to solve the known downstream tasks instead of learning a general representation. Thus, existing contrastive learning methods are still downstream-task dependent, as long as the data augmentation set $\mathcal{T}$ is manually selected.

\subsection{Generalized contrastive learning}
\label{sec5.2-general}

From the above discussion, we know that the contrastive learning methods are still dependent on downstream tasks. To solve this issue, researchers proposed several solutions to generalize contrastive learning.

\paragraph{General representation evaluation}

The reason why contrastive representation learning is biased towards the specific downstream tasks \cite{tian2020makes,xiao2020should} is because the standard evaluation of SSL is based on the downstream tasks. This standard evaluation not only induces biases, but also is highly abstract since it is only appropriate when the features are linearly separable. Thus, an intuitive solution is to evaluate the learned representation differently.

\begin{figure}
     \centering
     \begin{subfigure}[b]{0.49\textwidth}
         \centering
         \includegraphics[width=\textwidth]{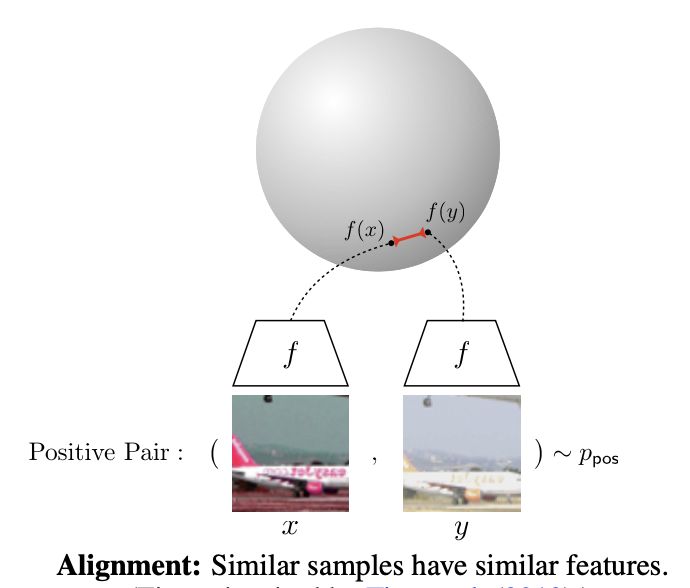}
     \end{subfigure}
     \hfill
     \begin{subfigure}[b]{0.49\textwidth}
         \centering
         \includegraphics[width=\textwidth]{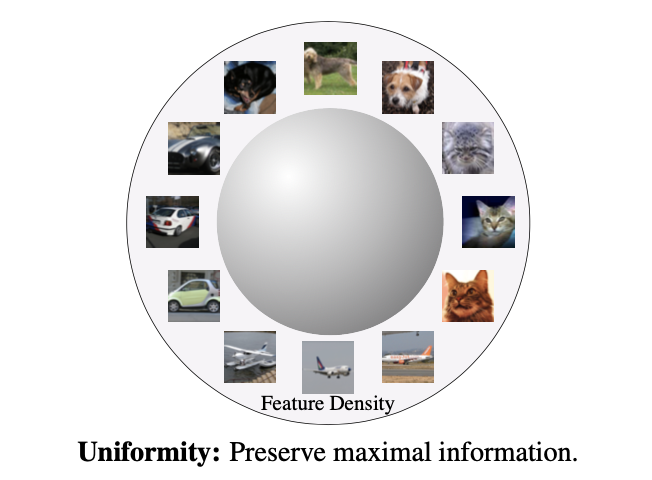}
     \end{subfigure}
    \caption{Two general representation evaluation methods, figures from \cite{wang2020understanding,tian2019contrastive}. Alignment measures the closeness between positive examples, uniformity measures the entropy of preserved information.}
    \label{fig: alignment-uniformity}
\end{figure}

Wang and Isola \cite{wang2020understanding} proposed to use two properties to measure the representation quality directly without the help of downstream tasks. They are 1) \emph{alignment}: the closeness between positive representation pairs, and 2) \emph{uniformity}: the amount of information preserved (entropy) via the induced distribution. Denote $p_{\text{pos}}$ as the positive data distribution and $p_{\text{data}}$ as the general data distribution, alignment and uniformity are mathematically defined as below based on the representation produced by a encoder $f_{\theta}$:
\begin{equation}
\begin{aligned}
    \mathcal{L}_{\text {align}}(f_\theta ; \alpha) & = \mathbb{E}_{(x, y) \sim p_{\text {pos }}}\left[\|f_\theta(x)-f_\theta(y)\|_{2}^{\alpha}\right] \\
    \mathcal{L}_{\text {uniform }}(f_\theta ; t) & = \log \mathbb{E}_{x, y \sim p_{\text {data }}}\left[e^{-t\|f_\theta(x)-f_\theta(y)\|_{2}^{2}}\right]
\end{aligned}
\end{equation}
where the alignment is the expected distance between positive pairs, and the uniformity measures if the data points follow a uniform distribution defined by the Gaussian potential kernel.
Defined as such, alignment and uniformity not only can measure the representation quality directly, but also provide a meaningful way for people to understand the learned representation.

Alignment and uniformity are validated by a set of theoretical and experimental analysis in \cite{wang2020understanding}. We summarize the key experiments as below: 1) the alignment and uniformity properties are statistically causally related with downstream task performances; 2) a direct optimization using the alignment and uniformity as a loss function can give comparable or better performances comparing with the performances obtained by optimizing the contrastive loss.
It can be inferred from the experimental results that the effective components of the contrastive loss may consist of the alignment and uniformity properties.

To conclude, the proposed evaluation metrics of the representation can debias and generalize representation learning because it is independent of the specific downstream tasks. As a follow-up work, \cite{chen2020intriguing} even showed that when the network is trained using $\mathcal{L}_{\text {align}}+\lambda \mathcal{L}_{\text {uniform}}$ as a loss function, the weight $\lambda$ is inversely related to the temperature scaling $\tau$ used in contrastive loss.

\paragraph{Keep covariance in pretext-invariant learning}

\begin{figure*}
\centering
    \includegraphics[width=\textwidth]{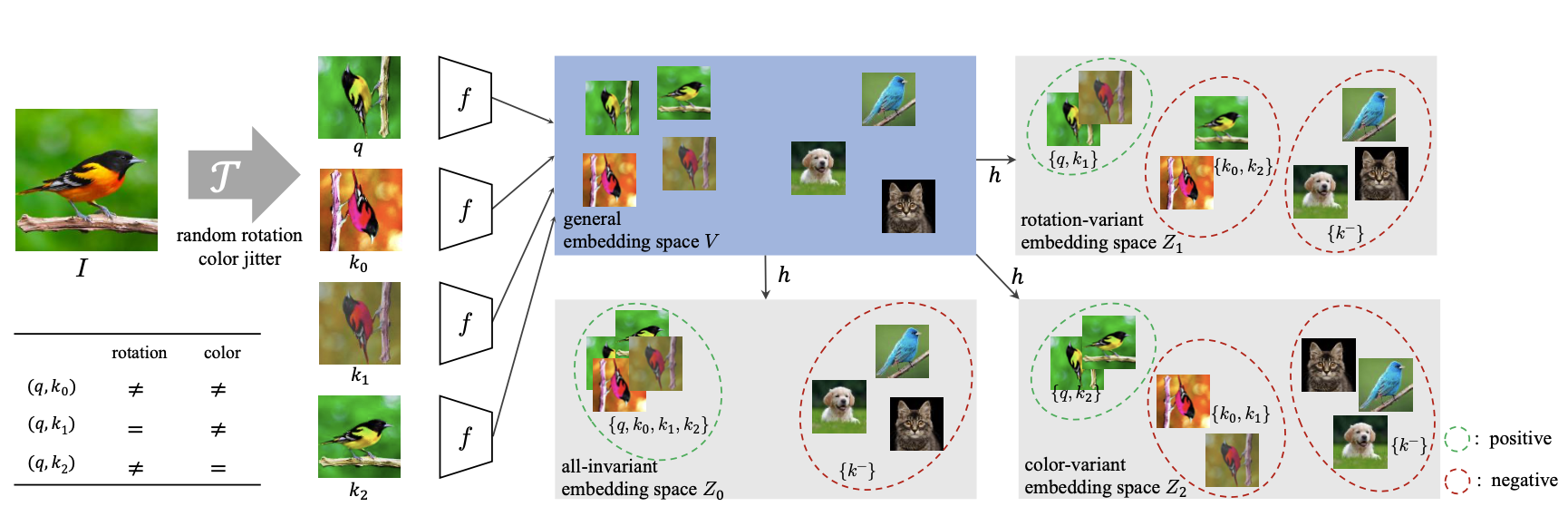}
    \caption{A method to keep covariance in pretext-invariant learning framework called LooC \cite{xiao2020should}. Images are augmented differently and passed into the encoder, where different embedding space preserves different information.}
    \label{fig-looc}
\end{figure*}

Another method to generalize the representation is to keep the covariance between features and data transformations when doing pretext-invariant learning. While learning features that are invariant with data augmentations sometimes helps the network to preserve visual simantics (section \ref{sec3-PIRL}), it discards data augmentation related information. However, the discarded information could be important when solving a different downstream task, e.g. a bird breed classification would require color information, which is typically discarded in pretext-invariant learning methods. Thus, to build a general representation, it is good to also keep covariance in the feature space.

$\dagger$ Leave-one-out contrastive learning (LooC) \cite{xiao2020should} \footnote{LooC is the main artifact in this section.} proposed a model that learns to capture both varying and invariant factors for representations by constructing multiple separate projector spaces, each of which is invariant to all but one augmentation (as shown in Fig. \ref{fig-looc}). By creating separate embedding spaces and concatenating the resulting representations, the final feature is aware of the augmentation-related information because each augmentation will be recognized by one feature space.
Denote $E_i^{+}$ as a positive logit and $E_i^{-}$ as a negative logit for the $i$-th embedding space $\mathcal{Z}_i$, where $i=0$ represents the original contrastive learning embedding space (all augmentation invariant) and $i=1, \ldots, k$ represents additional $i$-th augmentation covariant embedding space, where $k$ is the total amount of augmentations. The loss function of LooC is simply the summation of $k+1$ contrastive loss, as shown below:
\begin{equation}
    \mathcal{L}_{\text{LooC}}=-\frac{1}{n+1}\left(\log \frac{E_{0}^{+}}{E_{0}^{+}+\sum E_{0}^{-}}+\sum_{i=1}^{k} \log \frac{E_{i}^{+}}{E_{i}^{+}+ \sum E_{i}^{-}}\right)
\end{equation}
by learning this loss function, LooC learns to be covariant with different augmentations in different embedding spaces, which makes it generalize better when the downstream task is unknown.

\section{Conclusion}

This review analyzed the key components in contrastive learning methods. First, we introduce that pretext invariant learning methods \cite{misra2020self} outperform pretext covariant learning methods because the former preserves the mutual visual semantics under image transformation. After that, based on pretext invariant learning, contrastive learning methods e.g. SimCLR \cite{chen2020simple} used contrastive loss to achieve supervised-level representation learning performance, with a dependency on 1) large numbers of negative examples for contrasting; 2) a specific composition of data augmentation operations. In the following, we discuss works that tackle these two dependencies.
For negative examples, we introduced MoCHi \cite{kalantidis2020hard} as a possible method to understand and improve the selection of negative examples. For the specific composition of data augmentations, we discussed people's criticism \cite{tian2020makes} about it and introduced two possible solutions: a general representation evaluation method \cite{wang2020understanding} and a covariant preserved pretext invariant learning method \cite{xiao2020should}. With all these components being discussed, we hope this literature review could help people understand the contrastive learning and develop novel more methods to improve contrastive learning.

We refer readers to other existing comprehensive reviews as well. \cite{jing2020self} provides a comprehensive review of self-supervised learning methods for visual representation. \cite{jaiswal2021survey} recapitulates the journey of discovery of the contrastive learning methods. \cite{liu2020self} reviews self-supervised learning methods with a focus on generative methods and the comparison between generative methods and contrastive methods.

\bibliography{main}
\bibliographystyle{abbrv}

\end{document}